\documentclass{article}

\usepackage[nonatbib,final]{nips_2016_arxiv}

\pdfoutput=1
\usepackage[utf8]{inputenc} % allow utf-8 input
\usepackage[T1]{fontenc}    % use 8-bit T1 fonts
\usepackage{url}            % simple URL typesetting
\usepackage{graphicx} % more modern
\usepackage{booktabs}       % professional-quality tables
\usepackage{amsfonts}       % blackboard math symbols
\usepackage{nicefrac}       % compact symbols for 1/2, etc.
\usepackage{microtype}      % microtypography
\usepackage{floatrow}
\usepackage{amsmath}
\usepackage{amssymb}
\usepackage{algorithmic,algorithm}

\newcommand{\argmin}{\operatornamewithlimits{argmin}}
\newcommand{\x}{\mathbf{x}}
\newcommand{\z}{\mathbf{z}}

\newfloatcommand{capbtabbox}{table}[][\FBwidth]

\title{GANS for Sequences of Discrete Elements \\with the Gumbel-softmax Distribution}

\author{
    Matt Kusner\\
    Alan Turing Institute\\
    University of Warwick
    \And
    Jos\'e Miguel Hern\'andez-Lobato\\
    University of Cambridge
}

\begin{document}

\maketitle

\begin{abstract}
Generative Adversarial Networks (GAN) have limitations when the goal is to
generate sequences of discrete elements. The reason for this is that
samples from a distribution on discrete objects such as the multinomial are
not differentiable with respect to the distribution parameters. This problem
can be avoided by using the Gumbel-softmax distribution, which is a continuous
approximation to a multinomial distribution parameterized in terms of the
softmax function. In this work, we evaluate the performance of GANs based on
recurrent neural networks with Gumbel-softmax output distributions in the task
of generating sequences of discrete elements.
\end{abstract}

%!TEX=main.tex
\section{Introduction}
Generative adversarial networks (GANs) are methods for generating synthetic
data with similar statistical properties as the real one
\cite{goodfellow2014generative}. In the GAN methodology a discriminative neural network D is
trained to distinguish whether a given data instance is synthetic or real,
while a generative network G is jointly trained to confuse D by generating high
quality data. This approach has been very sucessful in computer vision tasks for
generating samples of natural images
\cite{denton2015deep,dosovitskiy2016generating,radford2016}.

GANs work by propagating gradients back from the discriminator D through the
generated samples to the generator G. This is perfectly feasible when the
generated data is continuous such as in the examples with images mentioned
above. However, a lot of data exists in the form of squences of discrete items.
For example, text sentences \cite{Bowman2016}, molecules encoded in the SMILE language \cite{gomez2016automatic}, etc. In these
cases, the discrete data is not differentiable and the backpropagated gradients
are always zero. 

Discrete data, encoded using a one-hot representation, can be sampled from a
multinomial distribution with probabilities given by the output of a softmax
function. The resulting sampling process is not differentiable.  However, we can obtain
a differentiable approximation by sampling from the Gumbel-softmax distribution
\cite{jang2016categorical}. This distribution has been previously used to train
variatoinal autoencoders with discrete latent variables \cite{jang2016categorical}. Here, we propose to
use it to train GANs on sequences of discrete tokens and we evaluate its
performance in this setting.

An alternative approach to train GANs on discrete sequences is described in
\cite{yu2016seqgan}. This method models the generation of the discrete sequence
as a stochastic policy in reinforcement learning and bypasses the generator
differentiation problem by directly performing gradient policy update.
% this is a tex file

\section{Gumbel-softmax distribution}
The softmax function can be used to parameterized a multinomial distribution
on a one-hot-encoding $d$-dimensional vector $\mathbf{y}$ in terms of a
continuous $d$-dimensional vector $\mathbf{h}$. Let $\mathbf{p}$ be a $d$-dimensional vector of probabilities
specifying the multinomial distribution on $\mathbf{y}$ with $p_i = p(y_i=1)$, $i=1,\ldots,d$. Then
\begin{align}
\mathbf{p} = \text{softmax}(\mathbf{h})\,\label{sec:gumbel:eq:0}
\end{align}
where $\text{softmax}(\cdot)$ returns here a $d$-dimensional vector with the output of the softmax function:
\begin{align}
\left[\text{softmax}(\mathbf{h})\right]_i = \frac{\exp(\mathbf{h}_i)}{\sum_{j=1}^K\exp(\mathbf{h}_j)}\,,\quad\text{for}\quad i = 1,\ldots,d\,. \label{sec:gumbel:eq:softmax_dim}
\end{align}
It can be shown that sampling $\mathbf{y}$ according to the previous multinomial distribution with probability vector 
given by (\ref{sec:gumbel:eq:0}) is the same as sampling $\mathbf{y}$ according to
\begin{align}
\mathbf{y} = \text{one\_hot}(\underset{i}{\arg\max} (h_i + g_i))\,,\label{sec:gumbel:eq:1}
\end{align}
where the $g_i$ are independent and follow a Gumbel distribution with zero location and unit scale.

The sample generated in (\ref{sec:gumbel:eq:1}) has gradient zero with respect to
$\mathbf{h}$ because the $\text{one\_hot}(\arg\max(\cdot))$
operator is not differentiable.
We propose to approximate this operator with a differentiable function based on the soft-max transformation \cite{jang2016categorical}.
In particular, we approximate $\mathbf{y}$ with 
\begin{align}
\mathbf{y} = \text{softmax}(1 / \tau (\mathbf{h} + \mathbf{g})))\,,\label{sec:gumbel:eq:2}
\end{align}
where $\tau$ is an inverse temperature parameter. When $\tau \rightarrow 0$, the samples generated by (\ref{sec:gumbel:eq:2})
have the same distribution as those generated by (\ref{sec:gumbel:eq:1}) and when $\tau \rightarrow \infty$,
the samples are always the uniform probability vector. For positive and finite values
of $\tau$ the samples generated by (\ref{sec:gumbel:eq:2}) are smooth and differentiable with respect to $\mathbf{h}$.

The probability distribution for (\ref{sec:gumbel:eq:2}), which is
parameterized by $\tau$ and $\mathbf{h}$, is called the Gumbel-softmax
distribution \cite{jang2016categorical}. A GAN on discrete data can then be
trained by using (\ref{sec:gumbel:eq:2}), starting with some relatively large
$\tau$ and then anealing it to zero during training.

\section{A recurrent neural network for discrete sequences}
In this section we describe how to construct a generative adversarial network (GAN) that is able to generate text from random noise samples. We also give a simple algorithm to train our model, inspired by recent work in adversarial modeling.

\begin{figure*}[t!]
\begin{center}
\centerline{\includegraphics[width=\textwidth,natwidth=787,natheight=415]{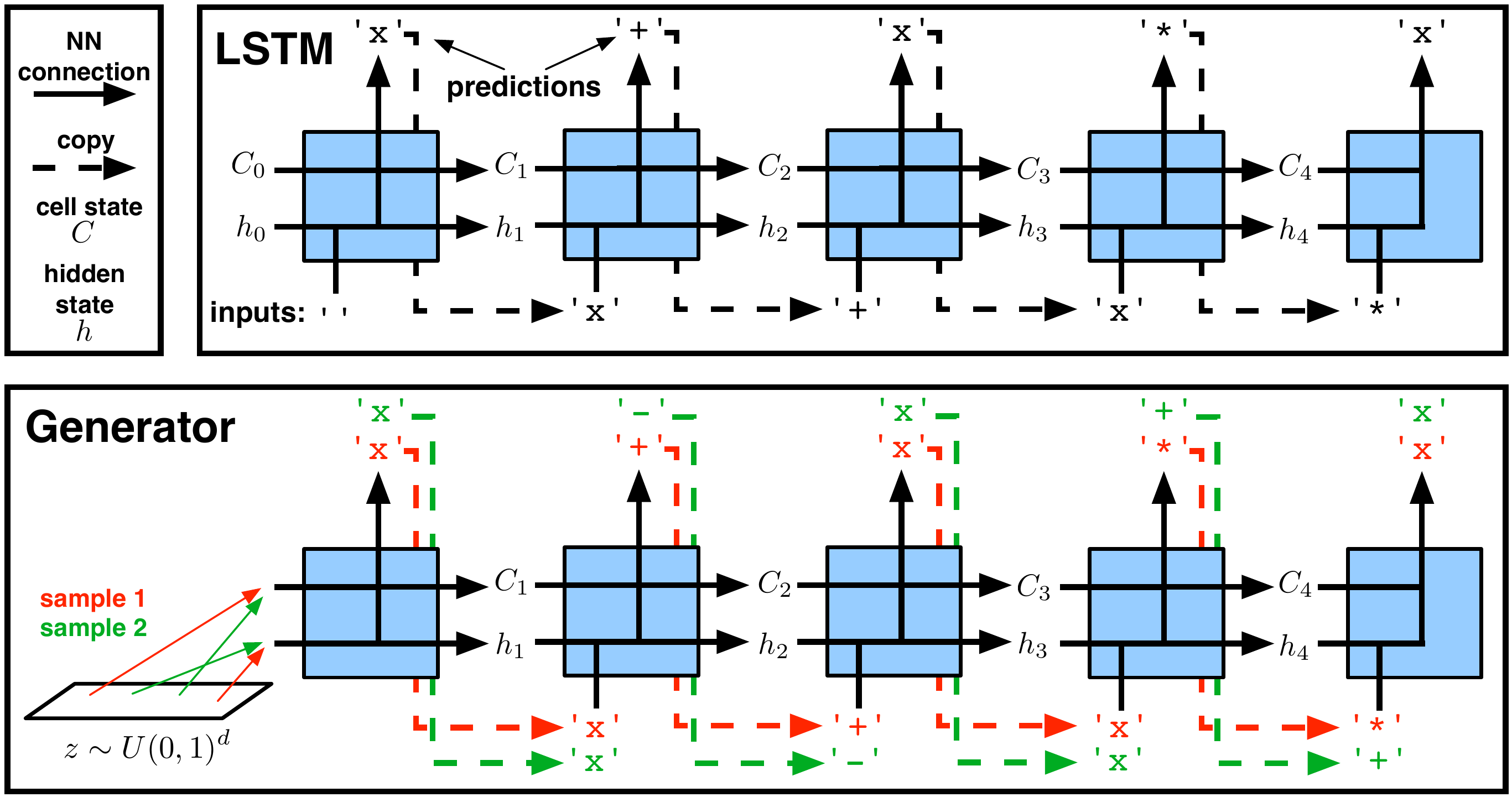}}
%\vspace{-2ex}
\caption{Models to generate simple one-variable arithmetic sequences. (\emph{Top}): The classic LSTM model during the prediction phase. Each LSTM unit (shown as a blue box) makes a prediction based on the input it as seen in the past. This prediction is then used as input to the next unit, which makes its own prediction, and so on. (\emph{Bottom}): Our generative model for discrete sequences. At the beginning we draw a pair of samples which are fed into the network in place of the initial cell state $C_0$ and hidden state $h_0$. Our trained network takes these samples and uses them to generate an initial character, this generated character is fed to the next cell in the LSTM as input, and so on.}
%\vspace{-5ex}
\label{figure.prediction}
\end{center}
\end{figure*}

\subsection*{An example}

Consider the problem of learning to generate simple one-variable arithmetic sequences that can be described by the following context-grammar:
\begin{align}
S \rightarrow \! x \! \mid \mid  \! S + S \! \mid \mid  \! S - S \! \mid \mid  \! S * S \! \mid \mid  \! S / S \! \nonumber
%S \rightarrow \! x \! & \! S \rightarrow S + S \! & \! S \rightarrow S - S \! & \! S \rightarrow S * S \! & \! S \rightarrow S / S \! \nonumber
\end{align}
where $\mid\mid$ divides possible productions of the grammar. The above grammar generates sequences of characters such as $x+x-x/x$ and $x-x*x*x*x$. 

Our generative model is based on a Long Short Term Memory (LSTM) recurrent neural network \cite{hochreiter1997long}, shown in the top of Figure~\ref{figure.prediction}. The LSTM is trained to predict a hidden-state vector $\mathbf{h}$ at every time-step (i.e., for every character). The softmax operator is then applied to $\mathbf{h}$ as in equations~(\ref{sec:gumbel:eq:softmax_dim}) and (\ref{sec:gumbel:eq:0}), whic gives a distribution over all possible generated characters (i.e., $x,+,-,/,*$). After training, the network generates data by sampling from the softmax distribution at each time-step.

One way to train the LSTM model to predict future characters is by matching the softmax distribution to a one-hot encoding of the input data via maximum likelihood estimation (MLE). In this work, we are interested in constructing a generative model for discrete sequences, which we will accomplish by sampling through the LSTM, as shown in the bottom of Figure~\ref{figure.prediction}. Our generative model takes as input a sample-pair which effectively replace the initial cell and hidden states. From this sample our generator constructs a sequence by successively feeding its predictions as input to the following LSTM unit. Our primary contribution is designing a method to train this generator to generate real-looking discrete sequences.

\subsection*{Generative adversarial modeling}

Given a set of $n$ data points $\{ \x_1, \x_2 \ldots, \x_n\}$ independently and identically drawn from a $d$-dimensional distribution $p(\x)$ (in our case each $\x$ is a one-hot encoding of a character), the goal of generative modeling is to learn a distribution $q(\x)$ that accuratley approximates $p(\x)$. The framework of generative adversarial modeling has been shown to yield models $q(\x)$ that generate amazingly realistic data points. The adversarial training idea is straight-forward. First, we are going to learn a so-called \emph{generator} $G$ that transforms samples from a simple, known distribution (e.g., a uniform or Gaussian distribution) into samples that approximate those drawn from $p(\x)$. Specifically, we define $q(\x) := G(\z)$, where $\z \sim U(0,1)^d$ (let $U(0,1)^d$ be the $d$-dimensional uniform distribution on the interval $[0,1]$). Second, to learn $G$ we will introduce a classifier we call the \emph{discriminator} $D$. The discriminator takes as input any real $d$-dimensional vector (this could be a generated input $G(\z)$ or a real one $\x$) and predicts the probability that the input is actually drawn from the real distribution $p(\x)$. It will be trained to take samples $G(\z)$ and real inputs $\x$ and accurately distinguish them. At the same time, the generator $G$ is trained so that it can fool the discriminator $D$ into thinking that a fake point it generated is real with high probability. Initially, the discriminator will be able to easily tell the fake points from the real ones and the generator is poor. However, as training progresses the generator uses this signal from the discriminator to determine how to generate more realistic samples. Eventually, the generator will generate samples so real that the discriminator will have a random chance of guessing if a generated point is real.

\subsection*{Using the Gumbel-softmax distribution}
In our case $G$ and $D$ are both LSTMs with parameters $\Theta$ and $\Phi$, respectfully. Our aim is to learn $G$ and $D$ by sampling inputs $\x$ and generated points $\z$, and minimizing differentiable loss functions for $G$ and $D$ to update $\Theta$ and $\Phi$. Unfortunately, sampling generated points $\z$ from the softmax distribution given by the LSTM, eq.~(\ref{sec:gumbel:eq:0}), is not differentiable with respect to the hidden states $\mathbf{h}$ (and thus $\Theta$). However, the Gumbel-softmax distribution, eq.~(\ref{sec:gumbel:eq:2}) is. Equipped with this trick we can take any differentiable loss function and optimize $\Theta$ and $\Phi$ using gradient-based techniques. We describe our adversarial training procedure in Algorithm~\ref{alg}, inspired by recent work on GANs \cite{sonderby2016amortised}. This algorithm can be shown in expectation to minimize the KL-divergence between $q(\z) \!=\! G(\z)$ and $p(\x)$.

%\begin{wrapfigure}{R}{0.45\textwidth}
%\vspace{-5ex}
% \begin{minipage}{0.45\textwidth}
\begin{algorithm}[H]                      % enter the algorithm environment
\caption{Generative Adversarial Network \cite{sonderby2016amortised}}          % give the algorithm a caption
\label{alg}                           % and a label for \ref{} commands later in the document
\begin{algorithmic}[1]                    % enter the algorithmic environment
%	\STATE \textbf{Input:} $\Vc$; $\Lambda \subseteq \mathbb{R}^d$; $T$; $(\epsilon,\delta)$; $\sigma^2_{\Vc,0}$; $\gamma_T$
%	\STATE $\mu_{\Vc,0} = 0$
	\STATE \textbf{data:} $\{ \x_1, \ldots, \x_n\} \sim p(\x)$,
	\STATE Generative LSTM network $G_\Theta$
	\STATE Discriminative LSTM network $D_\Phi$
	\WHILE{ loop until convergence }
		\STATE Sample mini-batch of inputs $B = \{\x_{B_1}, \ldots, \x_{B_m} \}$
		\STATE Sample noise $N = \{\z_{N_1}, \ldots, \z_{N_m}\}$
		\STATE Update discriminator $\Phi = \argmin_\Phi -\frac{1}{m} \sum_{\x \in B} \log D_\Phi(\x) - \frac{1}{m} \sum_{\z \in N} \log(1 - D_\Phi(G_\Theta(\z)))$
		\STATE Update generator $\Theta = \argmin_\Theta - \frac{1}{m} \sum_{\z \in N} \log \frac{D_\Phi(G_\Theta(\z))}{1-D_\Phi(G_\Theta(\z))}$
	\ENDWHILE
%	\STATE \textbf{Return:} $\tilde{\lambda},\tilde{v}$
\end{algorithmic}
\end{algorithm}
% \end{minipage}
%\vspace{-4ex}
%\end{wrapfigure}

\begin{figure*}[t!]
\begin{center}
\centerline{\includegraphics[width=\textwidth,natwidth=798,natheight=325]{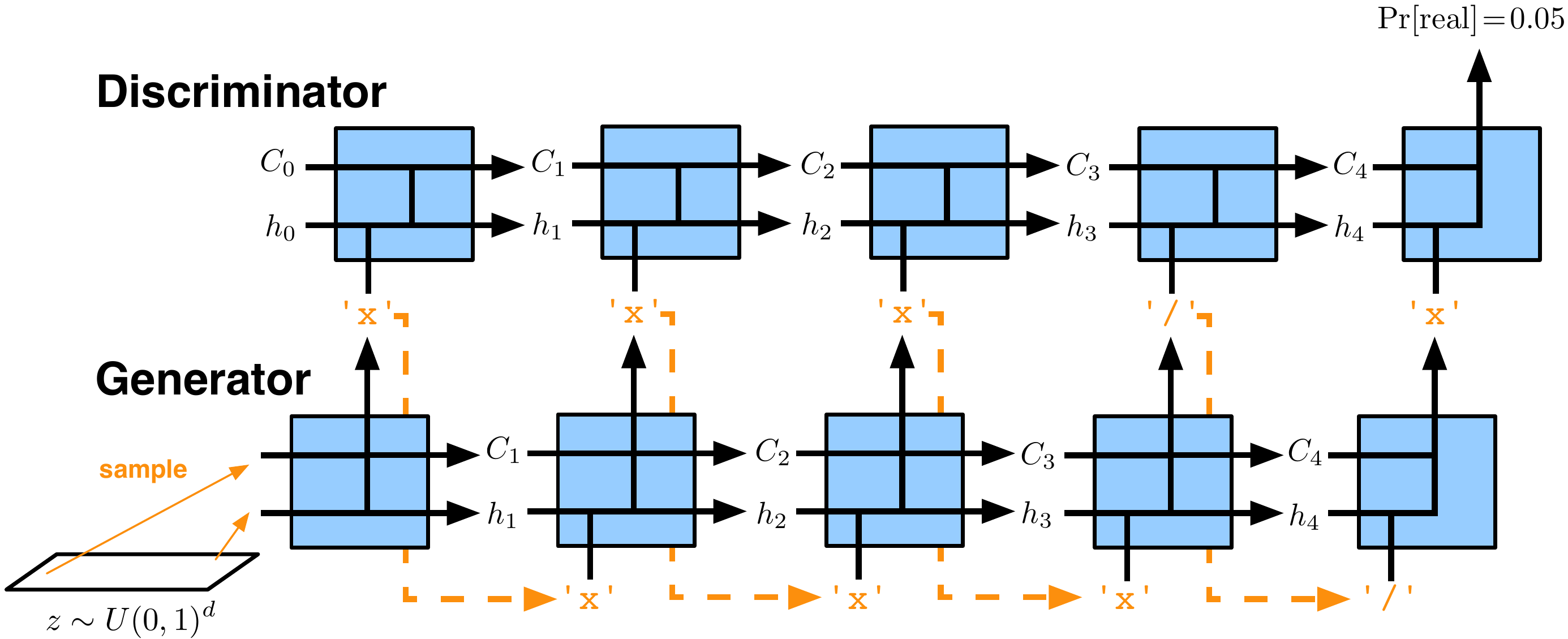}}
%\vspace{-2ex}
\caption{The adversarial training procedure. Our generative model first generates a full-length sequence. This sequence is fed to the discriminator (also a LSTM), which predicts the probability of it being a real sequence. Additionally (not shown), the discriminator is fed real discrete sequence data, which again it predicts the probability of it being real. The weights the networks are modified to make the discriminator better at recognizing real from fake data, and to make the generator better at fooling the discriminator.}
%\vspace{-5ex}
\label{figure.adversarial}
\end{center}
\end{figure*}

Figure~\ref{figure.adversarial} shows a schematic of the adversarial training procedure for discrete sequences.

\section{Experiments}
We now show the power of our adversarial modeling framework for generating discrete sequences. To illustrate this we consider modeling the context-free grammar introduced in Section 3. We generate $5000$ samples with a maximum length of $12$ characters from the context-free grammar (CFG) for our training set. We pad all sequences with less than $12$ characters with spaces.

\subsection*{Optimization details}
We train both the discriminator and generator using ADAM \cite{kingma2014adam} with a fixed learning rate of $0.001$ and a mini-batch size of $m\!=\!200$. Inspired by the work of \cite{sonderby2016amortised} who use input noise to stabilize GAN training, for every input $\x$ we form a vector $\mathbf{h}$ such that its softmax (instead of being one-hot) places a probability of approximately $0.9$ on the correct character and a probability of $(1-0.9)/(d-1)$ on the remaining $d\!-\!1$ characters. We then apply the Gumbel-softmax trick to generate a vector $\mathbf{y}$ as in equation~(\ref{sec:gumbel:eq:2}). We use this vector instead of $\x$ throughout training. We train the generator and discriminator for $20,000$ mini-batch iterations. During the training we linearly anneal the temperature of the Gumbel-softmax distribution, from $\tau\!=\!5$ (i.e., a very flat distribution) to $\tau\!=\!1$ (a more peaked distribution) for iterations $1$ to $10,000$ and then kept at $\tau\!=\!1$ until training ends.

\begin{figure*}[t!]
\begin{center}
\centerline{\includegraphics[width=\textwidth,natwidth=2208,natheight=468]{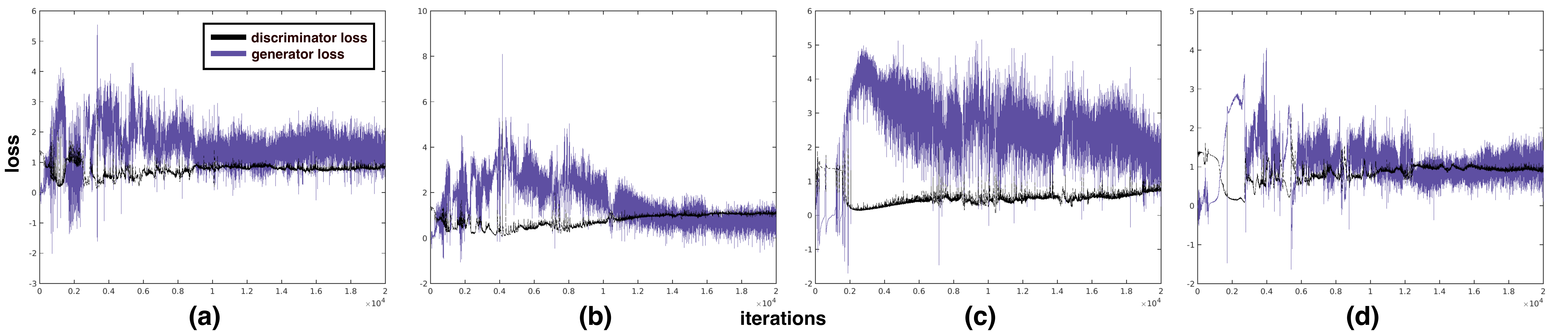}}
\vspace{-2ex}
\caption{The generative and discriminative losses throughout training. Ideally the loss of the discriminator should increase while the generator should decrease as the generator becomes better at mimicking the real data. \textbf{(a)} The default network with Gumbel-softmax temperature annealing. \textbf{(b)} The same setting as (a) but increasing the size of the generated samples to $1,000$. \textbf{(c)} Only varying the input vector temperature. \textbf{(d)} Only introducing random noise into the hidden state and not the cell state.}
%\vspace{-5ex}
\label{figure.losses}
\end{center}
\end{figure*}

\subsection*{Learning a CFG}
Figure~\ref{figure.losses} \textbf{(a)} shows the generator and discriminator losses throughout training for this setting. We experimented with increasing the size of the generated samples to $1,000$, as this has been reported to improve GAN modeling \cite{huszar2015not}, shown in Figure~\ref{figure.losses} \textbf{(b)}. We also experimented with just varying the temperature for the input vectors $\mathbf{y}$ and fixing the generator temperature to $\tau\!=\!1$ (in Figure~\ref{figure.losses} \textbf{(c)}).  Finally, we also tried just introducing random noise into the hidden state and allowing the network to learn an initial cell state $C_0$ (Figure~\ref{figure.losses} \textbf{(d)}).

\begin{figure*}[t!]
\begin{center}
\centerline{\includegraphics[width=\textwidth,natwidth=1139,natheight=735]{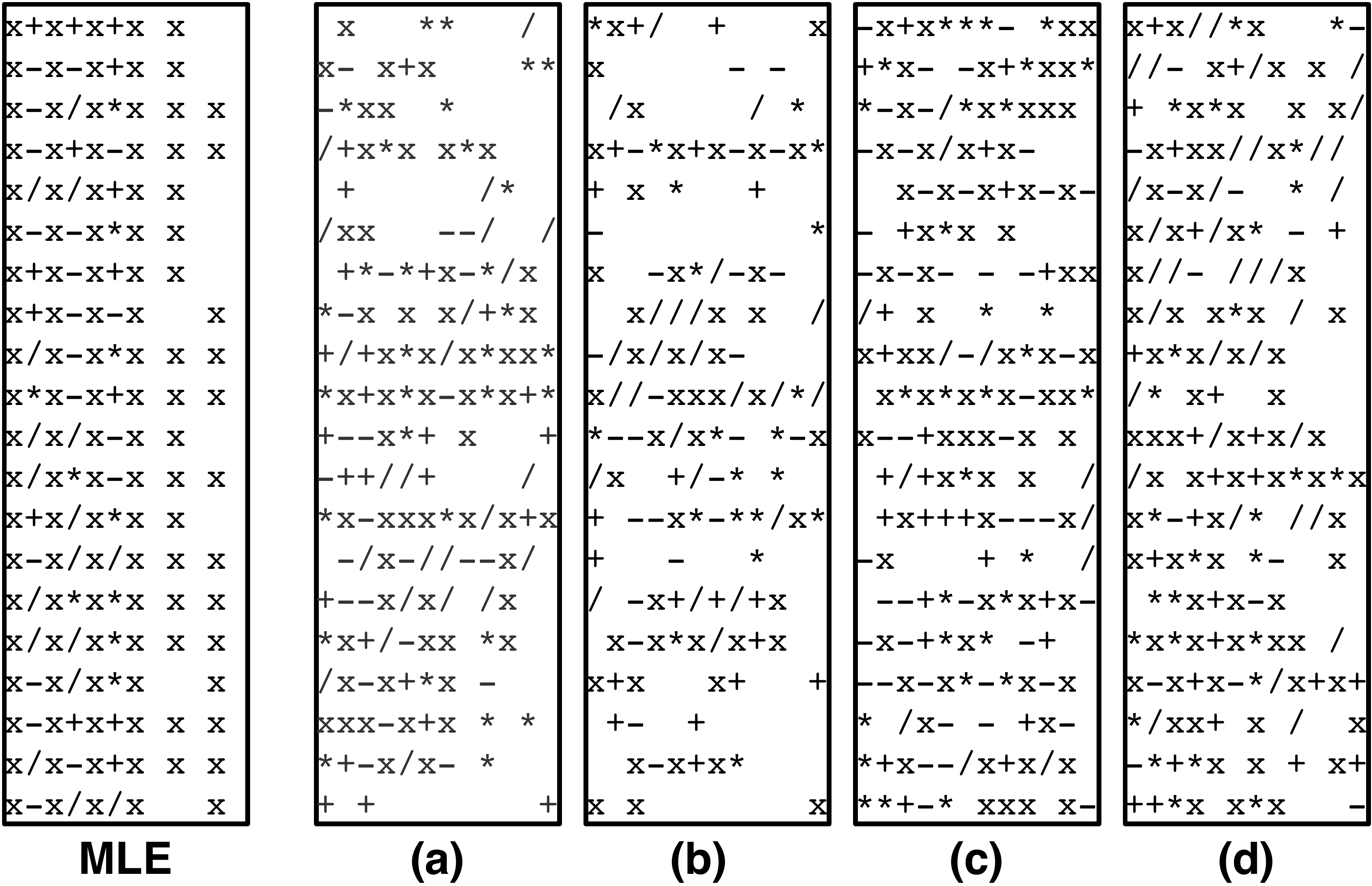}}
\vspace{-2ex}
\caption{The generated text for MLE and GAN models. The plots \textbf{(a)}-\textbf{(d)} correspond to the models of Figure~\ref{figure.losses}.}
%\vspace{-5ex}
\label{figure.generate}
\end{center}
\end{figure*}

Figure~\ref{figure.generate} shows the text generated by MLE and GAN models. Each row is a sample from either model, each consisting of $12$ characters (we have included the blank space character as some training inputs are padded with spaces if less than $12$ characters). While the MLE LSTM is not strictly a generative model in the sense of drawing a discrete sequence from a distribution, we include it for reference. We can see that our GAN models are learning to generate alternating sequences of $x$'s, similar to the MLE result. Specifically, the 4th, 10th, and 17th rows of plot \textbf{(a)}, show samples that are very close to the training data, and many such examples exist for the remaining plots as well. 

We believe that these results, as a proof of concept, show strong promise for training GANs to generate discrete sequence data. Further, we believe that incorporating recent advances in GANs such as training GANs using variational divergence minimization \cite{nowozin2016f} or via density ratio estimation \cite{uehara2016generative} could yield further improvements. We aim to experiment with these in future work.

\bibliography{references}
\bibliographystyle{plain}

\end{document}